%% file: main.tex
\documentclass{article} 
\usepackage{iclr2026_conference,times}
\input{math_commands.tex}

\usepackage{hyperref}
\usepackage{url}
\usepackage{booktabs}
\usepackage{graphicx}

\title{Evaluating Performance Drift from Model Switching in Multi-Turn LLM Systems}

\author{
Raad Khraishi$^{1,2}$ \quad
Iman Zafar$^{1}$ \quad
Katie Myles$^{1}$ \quad
Greig A.\ Cowan$^{1}$ \\
\\
$^{1}$NatWest AI Research \\
$^{2}$University College London
}

\iclrfinalcopy 
\begin{document}
\maketitle

\begin{abstract}
Deployed multi-turn LLM systems routinely switch models mid-interaction due to upgrades, cross-provider routing, and fallbacks. Such handoffs create a context mismatch: the model generating later turns must condition on a dialogue prefix authored by a different model, potentially inducing silent performance drift. We introduce a switch-matrix benchmark that measures this effect by running a prefix model for early turns and a suffix model for the final turn, and comparing against the no-switch baseline using paired episode-level bootstrap confidence intervals. 
Across CoQA conversational QA and Multi-IF benchmarks, even a single-turn handoff yields prevalent and statistically significant, directional effects and may swing outcomes by $-8$ to $+13$ percentage points in Multi-IF strict success rate and $\pm 4$ absolute F1 on CoQA, comparable to the no-switch gap between common model tiers (e.g., GPT‑5‑nano vs GPT‑5‑mini).
We further find systematic compatibility patterns: some suffix models degrade under nearly any non-self dialogue history, while others improve under nearly any foreign prefix. 
To enable compressed handoff risk monitoring, we decompose switch-induced drift into per-model prefix influence and suffix susceptibility terms, accounting for $\mathord{\sim}70\%$ of variance across benchmarks.
These results position handoff robustness as an operational reliability dimension that single-model benchmarks miss, motivating explicit monitoring and handoff-aware mitigation in multi-turn systems.
\end{abstract}

\section{Introduction}

Large language models (LLMs) are increasingly deployed as interactive systems where users engage in multi-turn dialogues rather than single prompts, and performance depends on maintaining state, adhering to evolving constraints, and staying consistent as context grows \citep{han2025can}. Prior work has shown that accuracy and instruction-following can degrade across turns and that earlier context can strongly shape later behavior \citep{kwan2024mt,he2024multi,hankache2025evaluating}. Yet most evaluations still implicitly assume a fixed model throughout an interaction.

In production, the model behind a conversation can change mid-session due to upgrades, cross-provider routing, or fallbacks \citep{chen2023frugalgpt}, and even within a product line, updates can induce behavioral drift \citep{chen2024chatgpt}. Yet we lack direct measurements of what happens when a model must continue from a dialogue history authored by a different model. This handoff is a structured distribution shift: the suffix model conditions on a prefix generated by another model rather than by itself. As with embedding-model upgrades \citep{yoon2025embedding}, mismatched conventions (verbosity/format) and implicit commitments can propagate across turns, including under prompt-injection pressure \citep{chang2025chatinject}.

In this paper, we introduce a switch-matrix benchmark for multi-turn systems that measures handoff-induced drift when one model must continue another model’s conversation. Drift is computed via paired comparisons to a no-switch baseline, making the effect attributable to the handoff rather than episode variance. 
We evaluate on CoQA \citep{reddy2019coqa} and Multi-IF \citep{he2024multi,zhou2023instruction}, two automatically-scored multi-turn benchmarks that stress conversational grounding and cumulative constraint adherence, using a diverse set of LLMs from leading providers, including Anthropic, OpenAI, and Google.
Across both tasks, a single-turn handoff yields prevalent, statistically significant, and directional effects. In Multi-IF, higher-performing prefix models (higher no-switch score) can boost weaker suffixes by anchoring a compliant output protocol. In CoQA, drift persists even though the original text passage remains in the model's context, suggesting a bias toward inherited assistant state rather than missing evidence. To enable compressed handoff risk monitoring, we also show drift is largely explained by two per-model factors: prefix influence and suffix susceptibility.

We present, to our knowledge, the first cross-provider switch-matrix measurement study that isolates handoff-induced drift in multi-turn LLM systems via paired comparisons to a no-switch baseline. Our contributions are: (1) we formalize model switching as an operational source of drift in multi-turn LLM systems and introduce a switch-matrix protocol to measure it relative to no-switch baselines; (2) we provide an efficient evaluation harness with prefix caching and paired episode-level bootstrap analysis; (3) we report cross-model, cross-provider switch matrices on CoQA and Multi-IF, showing that even final-turn switching can induce measurable drift not predicted by single-model benchmark scores; and (4) we decompose switch-induced drift into per-model prefix influence and suffix susceptibility terms, enabling compressed handoff-risk monitoring.

\section{Methodology}
\label{sec:method}

Let $\mathcal{M}=\{m_1,\dots,m_K\}$ be a set of LLMs and let an episode $e$ denote a multi-turn benchmark instance (a dataset row executed in a fixed environment). For each ordered model pair $(A,B)\in\mathcal{M}\times\mathcal{M}$ we run a context-switch cell $(A\!\rightarrow\!B)$: model $A$ generates the first $T$ assistant turns, then model $B$ generates the remaining turns until termination.
We focus on a final-turn switch policy, where $T=L-1$ and $L$ is the (fixed) number of assistant turns in the episode. This isolates a continuation problem: in every off-diagonal cell, the suffix model must produce a decisive final output while conditioning on a dialogue prefix authored by a different model, with the suffix responsible for exactly one turn across all cells.

For any suffix model $B$, the natural no-switch baseline is the diagonal cell $(B\!\rightarrow\!B)$, where $B$ authors the entire dialogue. We quantify the switch effect via paired per-episode differences,
\begin{equation*}
\delta_{A\rightarrow B}(e) \;=\; s_{A\rightarrow B}(e) - s_{B\rightarrow B}(e),
\end{equation*}
where $s(\cdot)$ is the episode score under the benchmark metric. We summarize switch effects with the episode mean
$\Delta_{A\rightarrow B}=\mathbb{E}_{e}[\delta_{A\rightarrow B}(e)]$; negative $\Delta_{A\rightarrow B}$ indicates that a prefix harms $B$ relative to the counterfactual where $B$ wrote its own context.

We evaluate switching on two deterministic, automatically-scored multi-turn benchmarks targeting complementary failure modes: conversational grounding and cumulative constraint adherence. As our protocol requires running a full $K\times K$ switch matrix over hundreds of episodes, yielding a large number of model calls, we prioritize benchmarks with fast, lightweight environments and inexpensive, fully automatic scoring to make the evaluation computationally and financially tractable.

\textbf{CoQA} \citep{reddy2019coqa} is conversational question answering over a passage.\footnote{\url{https://huggingface.co/datasets/stanfordnlp/coqa}} 
Each episode is a story with a sequence of questions whose answers depend on conversational state (e.g., coreference). We truncate each episode to the first $L=10$ question--answer turns. We require the model to output exactly one XML tag \texttt{<answer>...</answer>} per turn and use a robust parser fallback if the tag is missing. We score each turn using the standard token-overlap F1 (SQuAD/CoQA normalization) and report last-turn F1 under the final-turn switch policy.

\textbf{Multi-IF} \citep{he2024multi} is a 3-turn multilingual extension of IFEval \citep{zhou2023instruction}, where each turn adds verifiable constraints (formatting, required keywords, casing, length, etc.).\footnote{\url{https://huggingface.co/datasets/facebook/Multi-IF}}  We use a verifier\footnote{\url{https://github.com/josejg/instruction_following_eval}} to compute instruction satisfaction and take the episode score as turn-3 conversation-level strict success (binary success of satisfying all accumulated constraints). As such, $\Delta_{A\rightarrow B}$ reflects both (a) how well the prefix model satisfies early-turn constraints and (b) how well the suffix continues the induced protocol.

To make the full $K\times K$ matrix tractable without changing the measured behavior, we cache prefix-model generations on disk keyed by (task, episode id, prefix model) and reuse them across all suffix models. We run our experiments with temperature set to $0$ and 2048 maximum output tokens as well as reasoning effort and verbosity to low where supported. 

We report both: (i) absolute cell performance $\mathbb{E}_e[s_{A\rightarrow B}(e)]$ and (ii) switch effects $\Delta_{A\rightarrow B}$ relative to the no-switch baseline. To quantify uncertainty while respecting the paired design, we use bootstrap confidence intervals by resampling episodes with replacement. For cell means, we use BCa bootstrap CIs for $\mathbb{E}_e[s_{A\rightarrow B}(e)]$ computed from the set of episode scores in a cell. For switch effects, we use paired BCa bootstrap CIs for $\Delta_{A\rightarrow B}$, where per-episode differences $\delta_{A\rightarrow B}(e)$ are computed by pairing trials for an episode, ensuring that each bootstrap replicate preserves within-episode coupling between $(A\rightarrow B)$ and $(B\rightarrow B)$.

\section{Results}
\label{sec:results}

We run the $9\times 9$ prefix--suffix switch matrix on 200 randomly sampled episodes per benchmark using the final-turn handoff described in Section \ref{sec:method} and evaluate a diverse set of models from leading providers, including Anthropic, OpenAI, and Google.
We report cell means (reported in the Appendix) and switch effects $\Delta_{A\rightarrow B}$ relative to the no-switch baseline, using 1000 bootstrap resamples per cell (BCa for means; paired BCa for $\Delta$).

\begin{table*}[ht]
\vspace{-4mm}
\centering
\caption{CoQA switch effect relative to the no-switch baseline.
Each cell reports $\Delta_{A\rightarrow B}$, where rows represent prefix models $A$ and columns are suffix models $B$. Stars indicate bootstrapped CI excludes 0 at 90\% ($^{*}$), 95\% ($^{**}$), or 99\% ($^{***}$) confidence.} 
\label{tab:coqa_delta_vs_suffix}
\vspace{2mm}
\resizebox{\textwidth}{!}{%
\begin{tabular}{lrrrrrrrrr}
\toprule
 & gpt-5-nano-2025-08-07 & gpt-5-mini-2025-08-07 & gpt-5.2-2025-12-11 & gemini-3-flash-preview & gemini-2.5-flash & deepseek-v3.2 & qwen-2.5-72b-instruct & claude-haiku-4.5 & claude-sonnet-4.5 \\
\midrule
gpt-5-nano-2025-08-07 & 0.000\textsuperscript{\phantom{***}} & $-$0.005\textsuperscript{\phantom{***}} & 0.000\textsuperscript{\phantom{***}} & $-$0.021\textsuperscript{***} & $-$0.020\textsuperscript{**\phantom{*}} & $-$0.045\textsuperscript{***} & 0.002\textsuperscript{\phantom{***}} & 0.013\textsuperscript{\phantom{***}} & $-$0.008\textsuperscript{\phantom{***}} \\
gpt-5-mini-2025-08-07 & 0.013\textsuperscript{\phantom{***}} & 0.000\textsuperscript{\phantom{***}} & 0.007\textsuperscript{\phantom{***}} & 0.002\textsuperscript{\phantom{***}} & 0.016\textsuperscript{\phantom{***}} & $-$0.007\textsuperscript{\phantom{***}} & 0.013\textsuperscript{\phantom{***}} & 0.015\textsuperscript{\phantom{***}} & $-$0.002\textsuperscript{\phantom{***}} \\
gpt-5.2-2025-12-11 & 0.011\textsuperscript{\phantom{***}} & 0.004\textsuperscript{\phantom{***}} & 0.000\textsuperscript{\phantom{***}} & $-$0.006\textsuperscript{\phantom{***}} & 0.008\textsuperscript{\phantom{***}} & $-$0.016\textsuperscript{**\phantom{*}} & 0.027\textsuperscript{**\phantom{*}} & 0.017\textsuperscript{\phantom{***}} & $-$0.003\textsuperscript{\phantom{***}} \\
gemini-3-flash-preview & 0.042\textsuperscript{**\phantom{*}} & $-$0.004\textsuperscript{\phantom{***}} & 0.005\textsuperscript{\phantom{***}} & 0.000\textsuperscript{\phantom{***}} & 0.013\textsuperscript{**\phantom{*}} & $-$0.005\textsuperscript{\phantom{***}} & 0.030\textsuperscript{***} & 0.028\textsuperscript{**\phantom{*}} & 0.007\textsuperscript{\phantom{***}} \\
gemini-2.5-flash & $-$0.010\textsuperscript{\phantom{***}} & 0.007\textsuperscript{\phantom{***}} & 0.008\textsuperscript{\phantom{***}} & $-$0.015\textsuperscript{*\phantom{**}} & 0.000\textsuperscript{\phantom{***}} & $-$0.016\textsuperscript{\phantom{***}} & 0.029\textsuperscript{***} & 0.030\textsuperscript{***} & $-$0.002\textsuperscript{\phantom{***}} \\
deepseek-v3.2 & 0.006\textsuperscript{\phantom{***}} & $-$0.000\textsuperscript{\phantom{***}} & 0.008\textsuperscript{\phantom{***}} & $-$0.006\textsuperscript{\phantom{***}} & 0.005\textsuperscript{\phantom{***}} & 0.000\textsuperscript{\phantom{***}} & 0.025\textsuperscript{*\phantom{**}} & 0.026\textsuperscript{**\phantom{*}} & 0.004\textsuperscript{\phantom{***}} \\
qwen-2.5-72b-instruct & 0.014\textsuperscript{\phantom{***}} & $-$0.000\textsuperscript{\phantom{***}} & 0.002\textsuperscript{\phantom{***}} & $-$0.011\textsuperscript{**\phantom{*}} & 0.003\textsuperscript{\phantom{***}} & $-$0.019\textsuperscript{\phantom{***}} & 0.000\textsuperscript{\phantom{***}} & 0.020\textsuperscript{\phantom{***}} & $-$0.006\textsuperscript{\phantom{***}} \\
claude-haiku-4.5 & 0.012\textsuperscript{\phantom{***}} & $-$0.010\textsuperscript{\phantom{***}} & 0.004\textsuperscript{\phantom{***}} & $-$0.009\textsuperscript{\phantom{***}} & 0.008\textsuperscript{\phantom{***}} & $-$0.026\textsuperscript{*\phantom{**}} & 0.017\textsuperscript{\phantom{***}} & 0.000\textsuperscript{\phantom{***}} & 0.000\textsuperscript{\phantom{***}} \\
claude-sonnet-4.5 & 0.010\textsuperscript{\phantom{***}} & 0.013\textsuperscript{\phantom{***}} & 0.007\textsuperscript{\phantom{***}} & $-$0.010\textsuperscript{**\phantom{*}} & 0.010\textsuperscript{\phantom{***}} & $-$0.025\textsuperscript{**\phantom{*}} & 0.032\textsuperscript{***} & 0.024\textsuperscript{\phantom{***}} & 0.000\textsuperscript{\phantom{***}} \\
\bottomrule
\end{tabular}
}
\end{table*}

Our key finding is that switch-induced drift is directional and measurable even with a single suffix turn (see Tables~\ref{tab:coqa_delta_vs_suffix} and \ref{tab:multiif_delta_vs_suffix}).
On CoQA, $22\%$ ($16/72$) off-diagonal switches are significant at the 95\% level and on Multi-IF, $25\%$ ($18/72$) are significant.
Notably, GPT-5-nano leads to large changes in performance as both a prefix in CoQA and as a suffix in Multi-IF, illustrating that some models can be particularly influential as prefix generators and/or particularly sensitive as suffix continuations under context mismatch.
The effect is not symmetric: a handoff that harms $B$ when preceded by $A$ may be neutral or even beneficial in the reverse direction, implying that handoff robustness is a property of the ordered pair $(A,B)$ and the ``dialogue regime'' induced by the prefix, not simply a function of model quality.

Switch effects are also not uniformly distributed across suffix models.
On CoQA, DeepSeek-v3.2 shows the clearest fragility to heterogeneous prefixes (including multiple significant negative deltas relative to its no-switch baseline), whereas Qwen-2.5-72B and Claude-Haiku often improve under non-self prefixes (positive deltas in many rows; Table~\ref{tab:coqa_delta_vs_suffix}).
On Multi-IF, Gemini-2.5-flash is a striking outlier in the opposite direction: it improves under a wide range of foreign prefixes (many significant positive deltas), whereas Gemini-3-flash and Claude-Haiku exhibit sharper penalties for specific incompatible prefixes (Table~\ref{tab:multiif_delta_vs_suffix}).

In CoQA (Table~\ref{tab:coqa_delta_vs_suffix}), where the evidence remains in context, the suffix can still be ``mis-calibrated'' by the prefix.
Because the passage is included in the dialogue history, the suffix always has access to the source of truth even when it did not author earlier turns.
Consequently, switch effects here largely reflect a specific phenomenon: the suffix model sometimes treats prior assistant answers as conversational ``state'' (entity choices, coreference resolutions, implied commitments) and stays consistent with them instead of fully re-grounding on the passage.
We find the strongest degradation when switching from GPT-5-nano into DeepSeek-v3.2 ($\Delta\!\approx\!-0.04$), but also observe improvements (e.g., Gemini-3 $\rightarrow$ GPT-5-nano yields $\Delta\!\approx\!+0.04$), consistent with a stronger prefix model providing a better conversational scaffold for a weaker suffix.
In contrast, within-family switching among GPT-5 variants yields near-zero deltas, suggesting that shared conventions reduce mismatch at handoff.

\begin{table*}[th]
\vspace{-4mm}
\centering
\caption{Multi-IF switch effect relative to the no-switch baseline.
Each cell reports $\Delta_{A\rightarrow B}$, where rows represent prefix models $A$ and columns are suffix models $B$. Stars indicate bootstrapped CI excludes 0 at 90\% ($^{*}$), 95\% ($^{**}$), or 99\% ($^{***}$) confidence.}
\label{tab:multiif_delta_vs_suffix}
\vspace{2mm}
\resizebox{\textwidth}{!}{%
\begin{tabular}{lrrrrrrrrr}
\toprule
 & gpt-5-nano-2025-08-07 & gpt-5-mini-2025-08-07 & gpt-5.2-2025-12-11 & gemini-3-flash-preview & gemini-2.5-flash & deepseek-v3.2 & qwen-2.5-72b-instruct & claude-haiku-4.5 & claude-sonnet-4.5 \\
\midrule
gpt-5-nano-2025-08-07 & 0.000\textsuperscript{\phantom{***}} & $-$0.015\textsuperscript{\phantom{***}} & $-$0.025\textsuperscript{\phantom{***}} & $-$0.010\textsuperscript{\phantom{***}} & 0.045\textsuperscript{\phantom{***}} & 0.000\textsuperscript{\phantom{***}} & $-$0.010\textsuperscript{\phantom{***}} & $-$0.076\textsuperscript{**\phantom{*}} & $-$0.035\textsuperscript{\phantom{***}} \\
gpt-5-mini-2025-08-07 & 0.096\textsuperscript{***} & 0.000\textsuperscript{\phantom{***}} & $-$0.045\textsuperscript{*\phantom{**}} & $-$0.061\textsuperscript{***} & 0.051\textsuperscript{*\phantom{**}} & 0.005\textsuperscript{\phantom{***}} & 0.000\textsuperscript{\phantom{***}} & $-$0.045\textsuperscript{\phantom{***}} & $-$0.045\textsuperscript{\phantom{***}} \\
gpt-5.2-2025-12-11 & 0.106\textsuperscript{***} & 0.051\textsuperscript{**\phantom{*}} & 0.000\textsuperscript{\phantom{***}} & 0.005\textsuperscript{\phantom{***}} & 0.081\textsuperscript{***} & 0.020\textsuperscript{\phantom{***}} & 0.071\textsuperscript{***} & $-$0.015\textsuperscript{\phantom{***}} & $-$0.010\textsuperscript{\phantom{***}} \\
gemini-3-flash-preview & 0.086\textsuperscript{**\phantom{*}} & $-$0.005\textsuperscript{\phantom{***}} & $-$0.005\textsuperscript{\phantom{***}} & 0.000\textsuperscript{\phantom{***}} & 0.086\textsuperscript{***} & 0.040\textsuperscript{*\phantom{**}} & 0.061\textsuperscript{*\phantom{**}} & 0.015\textsuperscript{\phantom{***}} & $-$0.005\textsuperscript{\phantom{***}} \\
gemini-2.5-flash & 0.061\textsuperscript{*\phantom{**}} & 0.061\textsuperscript{**\phantom{*}} & 0.020\textsuperscript{\phantom{***}} & 0.010\textsuperscript{\phantom{***}} & 0.000\textsuperscript{\phantom{***}} & 0.015\textsuperscript{\phantom{***}} & 0.051\textsuperscript{\phantom{***}} & $-$0.015\textsuperscript{\phantom{***}} & $-$0.035\textsuperscript{\phantom{***}} \\
deepseek-v3.2 & 0.106\textsuperscript{***} & 0.015\textsuperscript{\phantom{***}} & $-$0.010\textsuperscript{\phantom{***}} & $-$0.005\textsuperscript{\phantom{***}} & 0.071\textsuperscript{***} & 0.000\textsuperscript{\phantom{***}} & 0.056\textsuperscript{*\phantom{**}} & 0.000\textsuperscript{\phantom{***}} & $-$0.035\textsuperscript{\phantom{***}} \\
qwen-2.5-72b-instruct & 0.121\textsuperscript{***} & 0.025\textsuperscript{\phantom{***}} & 0.020\textsuperscript{\phantom{***}} & $-$0.035\textsuperscript{\phantom{***}} & 0.061\textsuperscript{**\phantom{*}} & $-$0.015\textsuperscript{\phantom{***}} & 0.000\textsuperscript{\phantom{***}} & $-$0.040\textsuperscript{\phantom{***}} & $-$0.010\textsuperscript{\phantom{***}} \\
claude-haiku-4.5 & 0.096\textsuperscript{***} & 0.040\textsuperscript{\phantom{***}} & 0.035\textsuperscript{\phantom{***}} & $-$0.020\textsuperscript{\phantom{***}} & 0.066\textsuperscript{**\phantom{*}} & 0.000\textsuperscript{\phantom{***}} & 0.000\textsuperscript{\phantom{***}} & 0.000\textsuperscript{\phantom{***}} & $-$0.020\textsuperscript{\phantom{***}} \\
claude-sonnet-4.5 & 0.131\textsuperscript{***} & 0.005\textsuperscript{\phantom{***}} & 0.030\textsuperscript{\phantom{***}} & 0.025\textsuperscript{\phantom{***}} & 0.066\textsuperscript{**\phantom{*}} & 0.010\textsuperscript{\phantom{***}} & 0.035\textsuperscript{\phantom{***}} & 0.015\textsuperscript{\phantom{***}} & 0.000\textsuperscript{\phantom{***}} \\
\bottomrule
\end{tabular}

}
\end{table*}

In Multi-IF (Table~\ref{tab:multiif_delta_vs_suffix}), switching primarily perturbs protocol adherence and cumulative constraint tracking.
For Multi-IF we score turn-3 conversation-level strict success, making the benchmark highly sensitive to formatting and accumulated instruction adherence.
Here, the largest-magnitude effects are positive and suggest that stronger prefix models can substantially boost weaker suffixes by stabilizing a compliant output regime (e.g., Claude-Sonnet $\rightarrow$ GPT-5-nano improves success by $\sim$13 points).
Conversely, we also observe sharp degradations from specific cross-provider mismatches (e.g., GPT-5-mini $\rightarrow$ Gemini-3 decreases success by $\sim$6 points, and GPT-5-nano $\rightarrow$ Claude-Haiku decreases success by $\sim$7--8 points).
These patterns suggest that many switch-induced failures in Multi-IF arise from behavioral anchoring (e.g., the suffix model continuing (or failing to override) a formatting/constraint-following protocol induced by the prefix) rather than a loss of underlying capability.

\paragraph{Drift factorizes into prefix influence and suffix susceptibility.}
Beyond pairwise idiosyncrasies, we find that the mean switch-effect matrix exhibits a strong low-rank structure. We fit a two-way additive model on off-diagonal cells,
\begin{equation}
\Delta_{A\rightarrow B} \;=\; \mu + \alpha_A + \beta_B + \epsilon_{A,B},
\end{equation}
with constraints $\sum_A \alpha_A = 0$ and $\sum_B \beta_B = 0$. Here $\alpha_A$ captures the average prefix influence of model $A$ across suffix models, and $\beta_B$ captures the suffix susceptibility of model $B$ to non-self dialogue histories (positive values indicate that $B$ tends to improve under foreign prefixes).

Fitting this model to the off-diagonal switch effects explains $70\%$ of the variance on CoQA and $74\%$ on Multi-IF (LOO CV $R^2$). The largest-magnitude $\beta_B$ terms (see Appendix) align with the patterns in Tables~\ref{tab:coqa_delta_vs_suffix} and \ref{tab:multiif_delta_vs_suffix}: on CoQA, DeepSeek-v3.2 has strongly negative suffix susceptibility ($\beta\approx-0.024$), while Qwen-2.5-72B and Claude-Haiku-4.5 have positive susceptibilities ($\beta\approx+0.018$). On Multi-IF, GPT-5-nano and Gemini-2.5-flash have strongly positive susceptibilities ($\beta\approx+0.078$ and $\beta\approx+0.047$), while Claude-Sonnet-4.5 is strongly negative ($\beta\approx-0.042$). Across tasks, prefix influence factors are moderately correlated ($\rho\approx0.6$), whereas suffix susceptibility is less stable ($\rho\approx0.2$), suggesting prefix regimes transfer more consistently than continuation robustness.

\paragraph{Implications for model monitoring.} 
Switching introduces a discrete change point: the suffix model is evaluated on prefixes generated by the prefix model, not its own. Monitoring should therefore be switch-aware and log the authoring model per turn and directly monitor the first post-switch turn(s). Before upgrades or cross-provider fallbacks, run a ``handoff regression'' by replaying historical prefixes through candidate suffix models to estimate expected $\Delta_{A\rightarrow B}$ and flag risky pairs or identify potential uplifts. Alternatively, our factorization approach provides a compressed monitoring view and can be used to track per-model prefix influence and suffix susceptibility scores as predictors of handoff risk, and then prioritize deeper evaluation for pairs with large residuals.
These approaches may be used to gate routing or apply simple mitigations when deviations from the no-switch baseline are large (e.g., injecting a short handoff instruction). 

\section{Conclusion}
We introduced a switch-matrix protocol to quantify drift induced when a suffix LLM must continue from a dialogue prefix authored by a different model. Across CoQA and Multi-IF, even a final-turn-only handoff yields statistically significant and highly directional effects relative to running the suffix model from the start. Notably, some switches improve performance, indicating that compatibility depends on the behavioral regime established in the prefix (e.g., formatting and constraint-tracking) rather than missing evidence alone. These findings suggest that mid-session model changes should be treated as a first-class source of operational drift, warranting handoff-aware monitoring beyond per-model averages. Future work will extend to earlier and multi-turn switches, broader task suites, and mitigation strategies such as explicit handoff summaries, learned lightweight adapters, and routing policies optimized for cross-model continuity.


\bibliography{iclr2026_conference}
\bibliographystyle{iclr2026_conference}

\appendix
\section{Appendix}

\begin{table*}[ht]
\centering
\caption{CoQA absolute performance by switch cell (mean with 95\% CI).
Each entry reports mean last-turn F1 with a 95\% BCa bootstrap CI for the switch cell $(A\!\rightarrow\!B)$ on CoQA under a final-turn handoff. Rows are prefix models $A$ and columns are suffix models $B$; the diagonal corresponds to the no-switch condition $(B\!\rightarrow\!B)$.}
\label{tab:coqa_mean_ci95}
\vspace{2mm}
\resizebox{\textwidth}{!}{%
\begin{tabular}{lrrrrrrrrr}
\toprule
 & gpt-5-nano-2025-08-07 & gpt-5-mini-2025-08-07 & gpt-5.2-2025-12-11 & gemini-3-flash-preview & gemini-2.5-flash & deepseek-v3.2 & qwen-2.5-72b-instruct & claude-haiku-4.5 & claude-sonnet-4.5 \\
\midrule
gpt-5-nano-2025-08-07 & 0.689 [0.636,0.738] & 0.751 [0.694,0.796] & 0.751 [0.700,0.801] & 0.765 [0.713,0.812] & 0.758 [0.709,0.802] & 0.730 [0.672,0.775] & 0.726 [0.674,0.773] & 0.737 [0.685,0.787] & 0.776 [0.725,0.821] \\
gpt-5-mini-2025-08-07 & 0.703 [0.645,0.754] & 0.757 [0.711,0.802] & 0.757 [0.704,0.803] & 0.788 [0.726,0.831] & 0.793 [0.743,0.834] & 0.768 [0.717,0.810] & 0.737 [0.680,0.786] & 0.739 [0.686,0.782] & 0.783 [0.734,0.833] \\
gpt-5.2-2025-12-11 & 0.700 [0.645,0.757] & 0.760 [0.704,0.808] & 0.750 [0.697,0.799] & 0.780 [0.731,0.827] & 0.786 [0.735,0.832] & 0.758 [0.703,0.805] & 0.751 [0.697,0.798] & 0.740 [0.692,0.787] & 0.781 [0.731,0.822] \\
gemini-3-flash-preview & 0.731 [0.675,0.779] & 0.753 [0.705,0.802] & 0.755 [0.709,0.804] & 0.786 [0.732,0.829] & 0.790 [0.740,0.834] & 0.769 [0.717,0.815] & 0.754 [0.702,0.800] & 0.752 [0.703,0.798] & 0.791 [0.744,0.839] \\
gemini-2.5-flash & 0.679 [0.620,0.735] & 0.763 [0.713,0.808] & 0.759 [0.709,0.803] & 0.771 [0.723,0.825] & 0.778 [0.732,0.822] & 0.758 [0.705,0.802] & 0.753 [0.696,0.798] & 0.754 [0.704,0.799] & 0.782 [0.733,0.826] \\
deepseek-v3.2 & 0.695 [0.636,0.747] & 0.756 [0.704,0.803] & 0.758 [0.710,0.803] & 0.781 [0.728,0.825] & 0.783 [0.738,0.825] & 0.774 [0.728,0.820] & 0.749 [0.696,0.797] & 0.750 [0.703,0.796] & 0.789 [0.742,0.833] \\
qwen-2.5-72b-instruct & 0.703 [0.649,0.753] & 0.757 [0.707,0.806] & 0.753 [0.703,0.800] & 0.775 [0.725,0.819] & 0.781 [0.730,0.827] & 0.755 [0.712,0.801] & 0.724 [0.675,0.773] & 0.744 [0.694,0.790] & 0.779 [0.730,0.819] \\
claude-haiku-4.5 & 0.701 [0.646,0.752] & 0.746 [0.691,0.789] & 0.754 [0.697,0.799] & 0.777 [0.731,0.822] & 0.785 [0.739,0.830] & 0.749 [0.698,0.794] & 0.741 [0.691,0.789] & 0.724 [0.671,0.765] & 0.785 [0.736,0.824] \\
claude-sonnet-4.5 & 0.699 [0.647,0.748] & 0.770 [0.712,0.811] & 0.757 [0.707,0.800] & 0.776 [0.721,0.821] & 0.787 [0.742,0.833] & 0.749 [0.699,0.792] & 0.756 [0.701,0.799] & 0.748 [0.690,0.794] & 0.784 [0.737,0.826] \\
\bottomrule
\end{tabular}
}
\end{table*}

\begin{table*}[ht]
\centering
\caption{Multi-IF absolute performance by switch cell (mean with 95\% CI).
Each entry reports mean turn-3 conversation-level strict success rate with a 95\% BCa bootstrap CI for the switch cell $(A\!\rightarrow\!B)$ on Multi-IF under a final-turn handoff. Rows are prefix models $A$ and columns are suffix models $B$; the diagonal corresponds to the no-switch condition $(B\!\rightarrow\!B)$.}
\label{tab:multiif_mean_ci95}
\vspace{2mm}
\resizebox{\textwidth}{!}{%
\begin{tabular}{lrrrrrrrrr}
\toprule
 & gpt-5-nano-2025-08-07 & gpt-5-mini-2025-08-07 & gpt-5.2-2025-12-11 & gemini-3-flash-preview & gemini-2.5-flash & deepseek-v3.2 & qwen-2.5-72b-instruct & claude-haiku-4.5 & claude-sonnet-4.5 \\
\midrule
gpt-5-nano-2025-08-07 & 0.333 [0.273,0.399] & 0.419 [0.354,0.490] & 0.447 [0.382,0.513] & 0.490 [0.424,0.562] & 0.404 [0.338,0.480] & 0.399 [0.338,0.470] & 0.333 [0.268,0.394] & 0.354 [0.283,0.419] & 0.429 [0.356,0.500] \\
gpt-5-mini-2025-08-07 & 0.429 [0.364,0.500] & 0.434 [0.364,0.505] & 0.429 [0.364,0.505] & 0.439 [0.374,0.510] & 0.409 [0.338,0.475] & 0.404 [0.333,0.470] & 0.343 [0.275,0.409] & 0.384 [0.318,0.444] & 0.419 [0.348,0.495] \\
gpt-5.2-2025-12-11 & 0.439 [0.374,0.515] & 0.485 [0.419,0.556] & 0.475 [0.408,0.550] & 0.505 [0.429,0.566] & 0.439 [0.374,0.510] & 0.419 [0.354,0.490] & 0.414 [0.354,0.488] & 0.414 [0.343,0.485] & 0.455 [0.384,0.520] \\
gemini-3-flash-preview & 0.419 [0.348,0.485] & 0.429 [0.359,0.500] & 0.470 [0.404,0.540] & 0.500 [0.429,0.571] & 0.444 [0.374,0.510] & 0.439 [0.374,0.505] & 0.404 [0.335,0.480] & 0.444 [0.369,0.510] & 0.460 [0.391,0.535] \\
gemini-2.5-flash & 0.394 [0.328,0.465] & 0.495 [0.429,0.561] & 0.495 [0.424,0.566] & 0.510 [0.444,0.581] & 0.359 [0.288,0.424] & 0.414 [0.348,0.485] & 0.394 [0.326,0.470] & 0.414 [0.354,0.480] & 0.429 [0.354,0.490] \\
deepseek-v3.2 & 0.439 [0.369,0.505] & 0.449 [0.384,0.520] & 0.465 [0.398,0.530] & 0.495 [0.424,0.561] & 0.429 [0.369,0.505] & 0.399 [0.338,0.480] & 0.399 [0.335,0.470] & 0.429 [0.362,0.495] & 0.429 [0.359,0.495] \\
qwen-2.5-72b-instruct & 0.455 [0.384,0.525] & 0.460 [0.394,0.535] & 0.495 [0.424,0.566] & 0.465 [0.394,0.535] & 0.419 [0.358,0.490] & 0.384 [0.318,0.455] & 0.343 [0.278,0.409] & 0.389 [0.318,0.460] & 0.455 [0.389,0.525] \\
claude-haiku-4.5 & 0.429 [0.359,0.495] & 0.475 [0.400,0.545] & 0.510 [0.439,0.578] & 0.480 [0.409,0.551] & 0.424 [0.359,0.490] & 0.399 [0.328,0.470] & 0.343 [0.277,0.414] & 0.429 [0.359,0.495] & 0.444 [0.379,0.515] \\
claude-sonnet-4.5 & 0.465 [0.399,0.535] & 0.439 [0.374,0.510] & 0.505 [0.439,0.571] & 0.525 [0.455,0.593] & 0.424 [0.354,0.491] & 0.409 [0.343,0.480] & 0.379 [0.313,0.449] & 0.444 [0.373,0.520] & 0.465 [0.399,0.530] \\
\bottomrule
\end{tabular}
}
\end{table*}

\begin{figure}[t]
  \centering
  \includegraphics[width=0.49\textwidth]{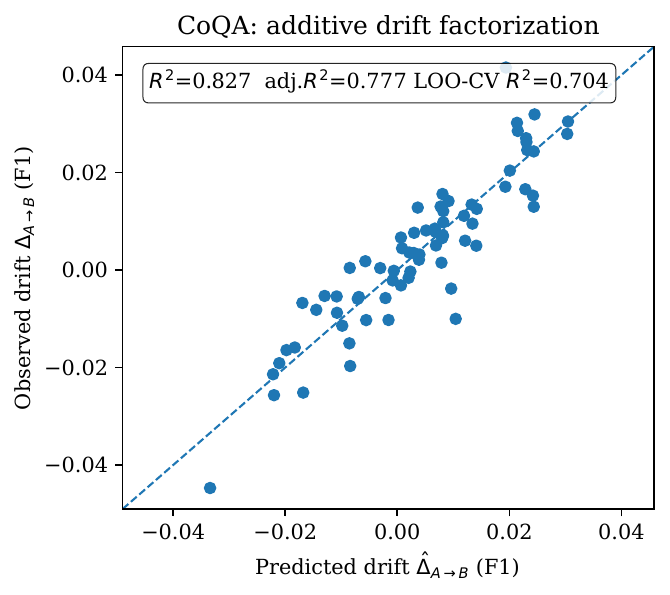}
  \includegraphics[width=0.49\textwidth]{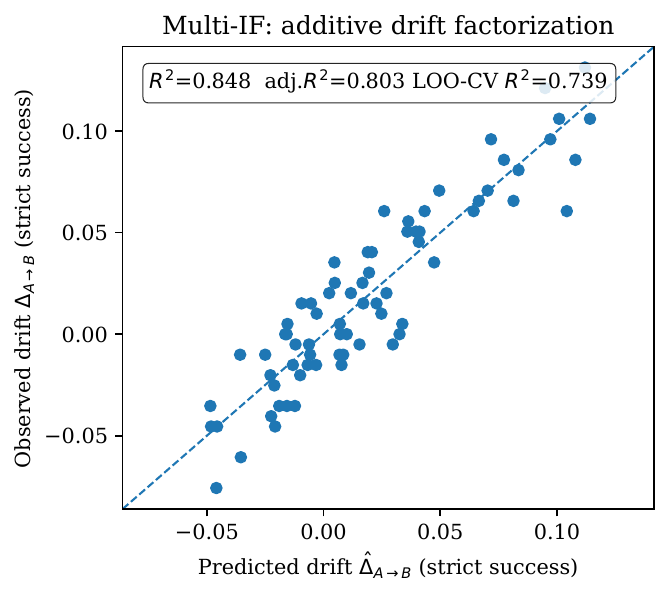}
  \caption{Observed vs.\ predicted mean drift under a two-way additive model $\hat\Delta_{A\to B}=c+\alpha_A+\beta_B$ on off-diagonal switch cells. CoQA (left): drift in last-turn F1. Multi-IF (right): drift in turn-3 strict success. The model explains most variance across switch cells ($R^2=0.83$ CoQA, $R^2=0.85$ Multi-IF), supporting a factorized view of handoff effects into prefix influence and suffix susceptibility.}
  \label{fig:factor_pred_vs_obs}
\end{figure}

\begin{figure}[t]
  \centering
  \includegraphics[width=0.49\textwidth]{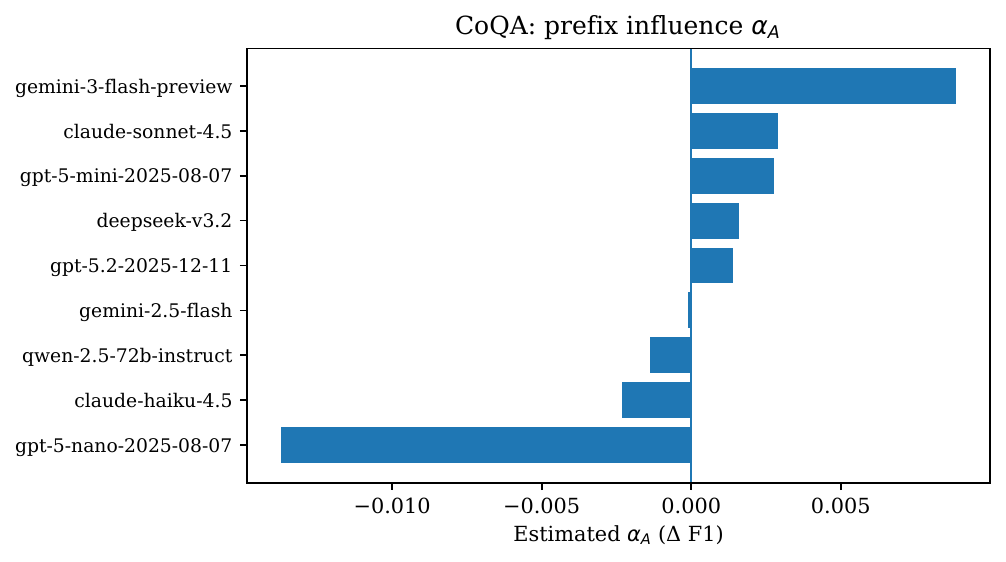}
 \includegraphics[width=0.49\textwidth]
{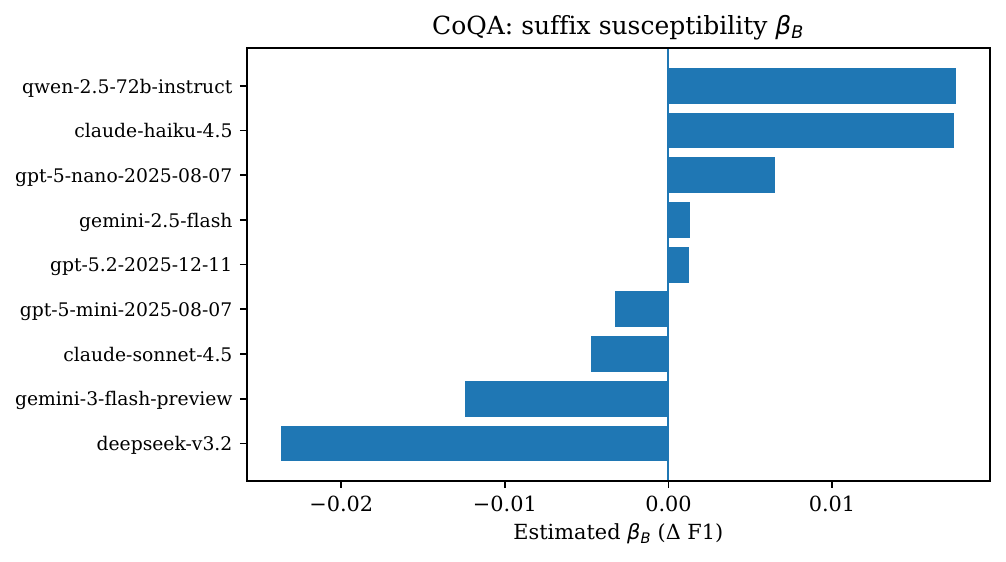}
  \caption{Estimated prefix influence $\alpha_A$ (sum-to-zero) and suffix susceptibility $\beta_B$ (sum-to-zero) from the additive drift model on CoQA. For $\alpha_A$, positive values indicate prefixes that tend to improve downstream suffix performance relative to suffix no-switch baselines; negative values indicate harmful prefix regimes. For $\beta_B$, positive values indicate suffix models that tend to benefit from foreign prefixes; negative values indicate degradation under context mismatch.}
  \label{fig:coqa_factor_alpha_beta}
\end{figure}

\begin{figure}[t]
  \centering
  \includegraphics[width=0.49\textwidth]{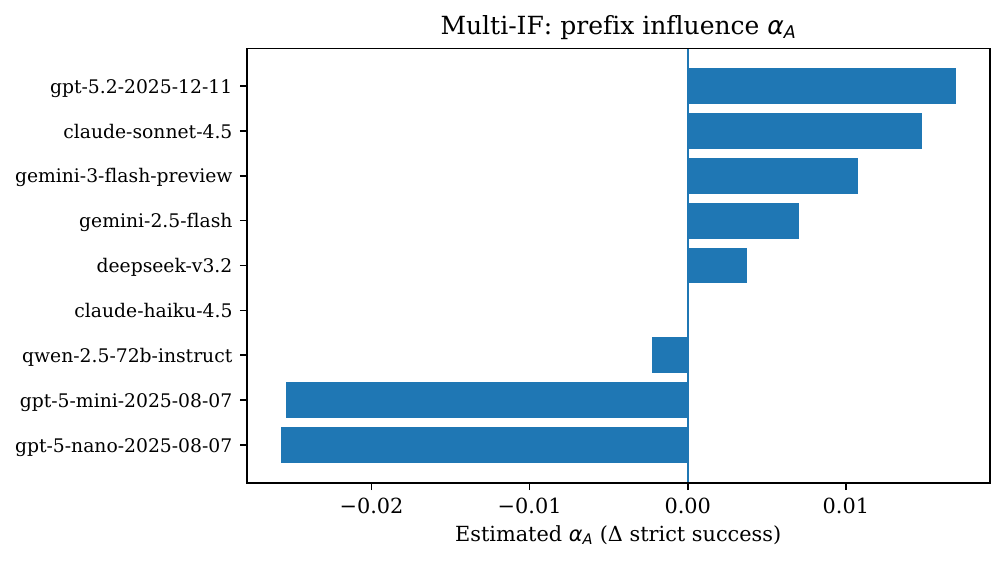}
 \includegraphics[width=0.49\textwidth]
{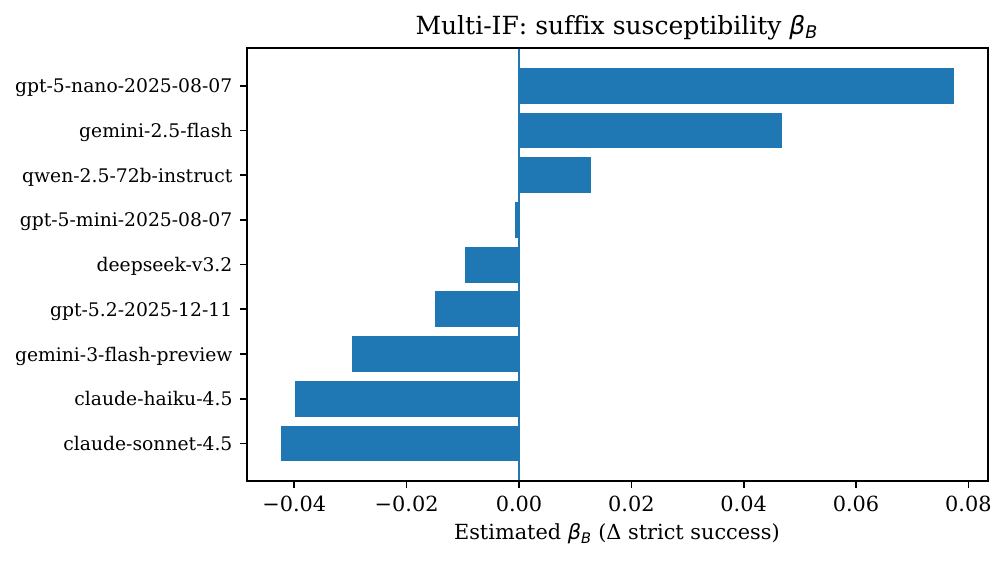}
  \caption{Estimated prefix influence $\alpha_A$ (sum-to-zero) and suffix susceptibility $\beta_B$ (sum-to-zero) from the additive drift model on Multi IF. For $\alpha_A$, positive values indicate prefixes that tend to improve downstream suffix performance relative to suffix no-switch baselines; negative values indicate harmful prefix regimes. For $\beta_B$, positive values indicate suffix models that tend to benefit from foreign prefixes; negative values indicate degradation under context mismatch.}
  \label{fig:multiif_factor_alpha_beta}
\end{figure}

\begin{table}[t]
\centering
\small
\caption{Largest pair-specific interaction residuals on CoQA after subtracting the additive prefix/suffix factorization (top 10 by $|\epsilon_{A,B}|$). Observed drift is the mean switch effect $\Delta_{A\to B}$ in last-turn F1; predicted drift is $\hat\Delta_{A\to B}=\mu+\alpha_A+\beta_B$ fit on off-diagonal cells.}
\resizebox{\linewidth}{!}{%
\begin{tabular}{llrrrr}
\toprule
Prefix $A$ & Suffix $B$ & Observed $\Delta_{A\to B}$ & Pred. $\hat\Delta$ & Resid. $\epsilon$ & $|\epsilon|$\\
\midrule
gemini-3-flash-preview & gpt-5-nano-2025-08-07 & $+$0.042 & $+$0.019 & $+$0.022 & 0.022\\
gemini-2.5-flash & gpt-5-nano-2025-08-07 & $-$0.010 & $+$0.010 & $-$0.020 & 0.020\\
gemini-3-flash-preview & gpt-5-mini-2025-08-07 & $-$0.004 & $+$0.010 & $-$0.013 & 0.013\\
gpt-5-mini-2025-08-07 & qwen-2.5-72b-instruct & $+$0.013 & $+$0.024 & $-$0.011 & 0.011\\
gpt-5-nano-2025-08-07 & deepseek-v3.2 & $-$0.045 & $-$0.033 & $-$0.011 & 0.011\\
gpt-5-nano-2025-08-07 & gemini-2.5-flash & $-$0.020 & $-$0.008 & $-$0.011 & 0.011\\
gpt-5-mini-2025-08-07 & deepseek-v3.2 & $-$0.007 & $-$0.017 & $+$0.010 & 0.010\\
claude-sonnet-4.5 & gpt-5-mini-2025-08-07 & $+$0.013 & $+$0.004 & $+$0.009 & 0.009\\
gemini-3-flash-preview & gpt-5.2-2025-12-11 & $+$0.005 & $+$0.014 & $-$0.009 & 0.009\\
gpt-5-mini-2025-08-07 & claude-haiku-4.5 & $+$0.015 & $+$0.024 & $-$0.009 & 0.009\\
\bottomrule
\end{tabular}}
\label{tab:coqa_top_residuals}
\end{table}

\begin{table}[t]
\centering
\small
\caption{Largest pair-specific interaction residuals on Multi-IF after subtracting the additive prefix/suffix factorization (top 10 by $|\epsilon_{A,B}|$). Observed drift is the mean switch effect $\Delta_{A\to B}$ in turn-3 strict success; predicted drift is $\hat\Delta_{A\to B}=\mu+\alpha_A+\beta_B$ fit on off-diagonal cells.}
\resizebox{\linewidth}{!}{%
\begin{tabular}{llrrrr}
\toprule
Prefix $A$ & Suffix $B$ & Observed $\Delta_{A\to B}$ & Pred. $\hat\Delta$ & Resid. $\epsilon$ & $|\epsilon|$\\
\midrule
gemini-2.5-flash & gpt-5-nano-2025-08-07 & $+$0.061 & +0.104 & $-$0.044 & 0.044\\
gemini-3-flash-preview & gpt-5-mini-2025-08-07 & $-$0.005 & $+$0.030 & $-$0.035 & 0.035\\
gemini-2.5-flash & gpt-5-mini-2025-08-07 & $+$0.061 & $+$0.026 & $+$0.035 & 0.035\\
claude-haiku-4.5 & qwen-2.5-72b-instruct & $+$0.000 & $+$0.033 & $-$0.033 & 0.033\\
claude-haiku-4.5 & gpt-5.2-2025-12-11 & $+$0.035 & $+$0.005 & $+$0.031 & 0.031\\
gpt-5-nano-2025-08-07 & claude-haiku-4.5 & $-$0.076 & $-$0.046 & $-$0.030 & 0.030\\
claude-sonnet-4.5 & gpt-5-mini-2025-08-07 & $+$0.005 & $+$0.034 & $-$0.029 & 0.029\\
qwen-2.5-72b-instruct & gpt-5-nano-2025-08-07 & +0.121 & $+$0.095 & $+$0.026 & 0.026\\
gpt-5-nano-2025-08-07 & gemini-3-flash-preview & $-$0.010 & $-$0.036 & $+$0.026 & 0.026\\
gpt-5-mini-2025-08-07 & gemini-3-flash-preview & $-$0.061 & $-$0.035 & $-$0.025 & 0.025\\
\bottomrule
\end{tabular}}
\label{tab:multi_if_top_residuals}
\end{table}

\end{document}

%% file: math_commands.tex
\usepackage{amsmath,amsfonts,bm}

\def\eqref#1{equation~\ref{#1}}

\def\1{\bm{1}}

\DeclareMathAlphabet{\mathsfit}{\encodingdefault}{\sfdefault}{m}{sl}
\SetMathAlphabet{\mathsfit}{bold}{\encodingdefault}{\sfdefault}{bx}{n}

